\documentclass[sigconf]{acmart}
\AtBeginDocument{%
  }

\setcopyright{acmlicensed}
\copyrightyear{2018}
\acmYear{2018}
\acmDOI{XXXXXXX.XXXXXXX}
\acmConference[Conference acronym 'XX]{Make sure to enter the correct
  conference title from your rights confirmation email}{June 03--05,
  2018}{Woodstock, NY}
\acmISBN{978-1-4503-XXXX-X/2018/06}

\usepackage{algorithm}
\usepackage{algorithmic}
\usepackage{caption}
\usepackage{float}
\usepackage{cuted}
\usepackage{stfloats}
\usepackage{makecell}
\usepackage{booktabs}
\usepackage{multirow}
\usepackage{graphicx}
\usepackage{subcaption}

\newcommand{\squishlist}{
	\begin{list}{$\bullet$}
		{ \setlength{\itemsep}{1pt}
			\setlength{\parsep}{1pt}
			\setlength{\topsep}{2.5pt}
			\setlength{\partopsep}{0.5pt}
			\setlength{\leftmargin}{1em}
			\setlength{\labelwidth}{1em}
			\setlength{\labelsep}{0.6em}
		}
	}
	\newcommand{\squishend}{
	\end{list}
}

\usepackage{xspace}
\newcommand{\method}{\textsc{MTServe}\xspace}

\usepackage{pifont}
\newcounter{squishcnt}

\newcommand{\squishlistcircled}{
	\begin{list}{\ding{\numexpr201+\value{squishcnt}\relax}}
		{ 
            \usecounter{squishcnt}
			\setlength{\itemsep}{1pt}
			\setlength{\parsep}{1pt}
			\setlength{\topsep}{2.5pt}
			\setlength{\partopsep}{0.5pt}
			\setlength{\leftmargin}{1em}
			\setlength{\labelwidth}{1em}
			\setlength{\labelsep}{0.6em}
		}
	}
\newcommand{\squishendcircled}{
	\end{list}
}

\newcommand{\stitle}[1]{\vspace*{0.5em}\noindent{\bf #1\/}}

\begin{document}

\title
[MTServe: Efficient Serving for Generative Recommendation Models with Hierarchical Caches]
{MTServe: Efficient Serving for Generative Recommendation \\
Models with Hierarchical Caches}

\author{
    Xin Wang$^{\dagger}$ \quad
    Chi Ma$^{\S}$ \quad
    Shaobin Chen$^{\dagger}$ \quad
    Pu Wang$^{\S}$ \quad
    Menglei Zhou$^{\S}$ \quad
    Junyi Qiu$^{\P}$ \\
    Qiaorui Chen$^{\P}$ \quad
    Jiayu Sun$^{\P}$ \quad
    Shijie Liu$^{\P}$ \quad
    Zehuan Wang$^{\P}$ \quad
    Lei Yu$^{\S}$ \quad
    Chuan Liu$^{\S}$ \quad
    Fei Jiang$^{\S}$ \\
    Wei Lin$^{\S}$ \quad
    Hao Wang$^{\dagger}$ \quad
    Jiawei Jiang$^{\dagger}$  \quad
    Xiao Yan$^{\dagger}$ \\
    $^{\dagger}$School of Computer Science, Wuhan University 
    \quad $^{\S}$Meituan 
    \quad $^{\P}$Nvidia
}


\renewcommand{\shortauthors}{Wang et al.}

\begin{abstract}
Generative recommendation (GR) offers superior modeling capabilities but suffers from prohibitive inference costs due to the repeated encoding of long user histories. 
While cross-request Key-Value (KV) cache reuse presents a significant optimization opportunity, the massive scale of individual user states creates a storage explosion that far exceeds physical GPU limits. 
We propose \method, a hierarchical cache management system that virtualizes GPU memory by leveraging host RAM as a scalable backup store. 
To bridge the I/O gap between tiers, \method introduces a suite of system-level optimizations, including a hybrid storage layout, an asynchronous data transfer pipeline, and a locality-driven replacement policy. 
On both public and production datasets, \method delivers up to 3.1$\times$ speedup while maintaining near-perfect hit ratios ($>98.5\%$). 

\end{abstract}





\received{20 February 2007}
\received[revised]{12 March 2009}
\received[accepted]{5 June 2009}

\maketitle

\section{Introduction}
\label{sec:introduction}

For years, recommendation systems have been dominated by traditional deep learning models, such as Wide \& Deep~\cite{cheng2016wide} and Deep Interest Networks~\cite{zhou2019deep}, 
which primarily focus on discriminative ranking through point-wise scoring. 
However, inspired by the transformative success of Large Language Models (LLMs) in capturing complex patterns, the industry is rapidly pivoting toward the paradigm of 
generative recommendation (GR)~\cite{zhai2024actions, deng2025onerec, chen2024hllm, han2025mtgr, huang2025towards, wang2025mtgenrecefficientdistributedtraining}. 
By framing user--item interaction modeling as a sequence transduction problem, 
GR models leverage Transformer-based architectures to unlock high-order dependencies within extensive interaction histories that were previously under-explored. 
While major industrial players are aggressively pursuing this paradigm to achieve unprecedented personalization, a rapidly evolving landscape we summarize in Appendix~\ref{app:related_work}, the transition has encountered a formidable obstacle: the prohibitive computational cost of processing long-term histories at scale. 
Consequently, satisfying the strict latency requirements of online services often demands an unsustainable amount of hardware resources,
making serving infrastructure optimization a critical priority for large-scale deployment.

\begin{figure}[!t]
    \centering
     \includegraphics[width=\linewidth]{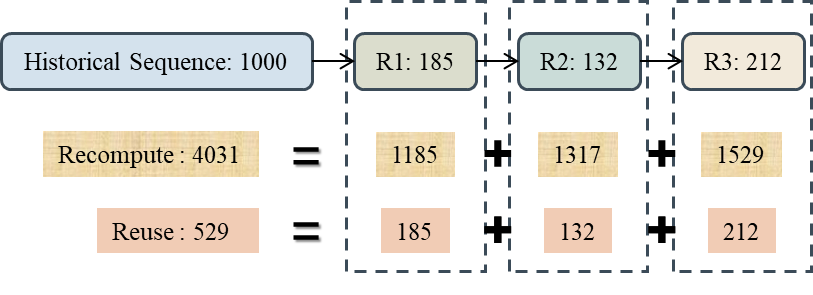} 
    \caption{Comparison of total tokens processed across three consecutive requests from the same user: \textit{Recompute} (4,031 tokens) vs. \textit{Reuse} (529 tokens).}
    \label{fig:motivation}
\end{figure}

\stitle{Cache Reuse and Challenges.} 
The inference workload in GR is characterized by processing a user's interaction sequence, which typically comprises an extensive historical prefix followed by a relatively small segment of new interactions. 
As illustrated in Figure~\ref{fig:motivation}, the incremental portion appended during each visit is often a tiny fraction of the total sequence length.
However, standard inference engines often treat each request as an isolated event, leading to massive and unnecessary computational redundancy. 
To mitigate this, \emph{user-side Key--Value (KV) caching} persists the intermediate activations—specifically the Key and Value tensors—of a specific user's historical prefix across multiple independent visits. 
By retrieving these user-level persistent states, the system can bypass the expensive re-encoding of the invariant history and focus exclusively on the incremental updates. 
As shown in Figure~\ref{fig:motivation}, for the three consecutive requests ($R_1, R_2, R_3$) from the same user, the recompute paradigm (w.o. reuse) is forced to repeatedly encode the entire cumulative prefix, totaling \emph{4,031 tokens}. 
In contrast, by leveraging user-side cache reuse, the computational workload is significantly reduced to merely \emph{529 tokens}.

Despite the potential computation reduction of KV cache reuse, persisting KV caches in GR at an industrial scale faces formidable challenges that render existing LLM-centric solutions inadequate.
\\
\textbf{\ding{172} Individual Persistence and Storage Explosion.} 
The primary hurdle stems from the sheer scale of the user base coupled with long interaction histories.
Unlike LLM serving where multiple users often share a common prefix (e.g., system prompts) that can be managed via radix trees~\cite{zheng2024sglang}, GR requires \emph{individualized persistence} for each unique user sequence. 
This leads to a prohibitive storage consumption: 
for a representative HSTU model~\cite{zhai2024actions}, serving just 1,000 users with 10,000-token histories would require over 160 GB of KV cache---already exceeding the total HBM capacity of two NVIDIA A100 (80GB) GPUs. 
With millions of active users, the memory demand far overwhelms any physical device limits.
\\
\textbf{\ding{173} Long-term Persistence vs. Session-based Caching.} 
Standard LLM caching~\cite{kwon2023efficient} is session-oriented and volatile, where KV states are discarded once a dialogue concludes. 
In contrast, GR requires \textit{long-term durability} across intermittent visits. Reactive policies (offloading only when memory is full) are insufficient here: they risk data loss if slots are reclaimed before being backed up, or cause severe latency spikes during forced synchronization. This necessitates a proactive management to asynchronously persist KV states to host memory before they become candidates for GPU reclamation.
\\
\textbf{\ding{174} High I/O-to-Computation Ratio and Frequent State Swapping.} 
The architectural and workload differences create a severe I/O bottleneck. 
While LLMs are compute-heavy and benefit from keeping KV states resident during long decoding sessions, recommendation models are relatively shallow and process independent, short-lived requests. 
This leads to \emph{frequent cache churn}, where almost every request triggers a mandatory restoration of a unique user state from host memory. 
Consequently, GR exhibits a much higher I/O-to-computation ratio, where the time to move data across the PCIe bus can easily exceed the model's execution time. 
Without careful orchestration, this constant state swapping will inevitably stall the inference pipeline, thereby negating the throughput benefits of caching and significantly increasing online inference latency.

\stitle{Our Solution \method.} 
To address these challenges, we propose \method, a hierarchical KV cache management system designed for serving generative recommendation in MeiTuan.
Developed in collaboration with NVIDIA to leverage high-performance hardware primitives, \method breaks the GPU memory wall by organizing the cache into a two-tier hierarchy. 
Instead of constraining user states solely to the capacity-limited GPU VRAM, \method utilizes the vast host memory as a scalable backup store while dynamically retaining frequently accessed states on the GPU. 
This architecture effectively virtualizes the device memory, providing the model with much larger cache capacity while significantly conserving expensive GPU resources.

To implement this tiered architecture efficiently, \method incorporates three key designs.
First, we introduce a \textit{dual-granularity storage abstraction} (Page-Chunk) to reconcile the conflict between computational flexibility and transmission efficiency. 
By coupling fine-grained GPU paging with coarse-grained CPU chunking, \method reduces GPU memory fragmentation while successfully saturating PCIe bandwidth. 
Second, to mitigate the high I/O-to-computation ratio, we design a \textit{latency-hiding transfer pipeline} that leverages double-buffered Direct Memory Access (DMA) and custom-optimized scatter/gather kernels. Through fine-grained layer-wise synchronization, \method effectively overlaps the mandatory I/O of state restoration with the model's forward pass, masking the transfer latency behind the computation. 
Finally, we employ a \textit{locality-driven management strategy} that exploits the temporal bursts of user requests. 
By adopting an LRU-based policy and decoupling data persistence from physical slot reclamation, \method enables zero-copy metadata eviction, ensuring that the critical inference path remains non-blocking even under heavy cache churn.

We implement \method using PyTorch and evaluate it on both the public KuaiRand-1K dataset and a production MT dataset. 
Compared to the recompute paradigm, \method achieves significant latency speedups of \textbf{3.04$\times$} and \textbf{3.1$\times$} at a batch size of 8, respectively. 
These improvements in latency directly translate into enhanced system throughput and a substantial reduction in computational resource consumption.
Furthermore, our system maintains a near-perfect total hit ratio ($>$\textbf{98}\%) even when the active working set significantly exceeds physical GPU memory capacity. 
These results validate \method's effectiveness and scalability for industrial-scale generative recommendation.

In summary, our main contributions are as follows:
\squishlist
    \item We identify the significant opportunity for user-side KV cache reuse in generative recommendation and characterize the core challenges—including individual storage explosion, long-term durability requirements, and high I/O-to-computation ratios—that render existing LLM-centric caching solutions inadequate.
    \item We propose \method, a hierarchical serving system that breaks the GPU memory wall by organizing the cache into a two-tier (GPU-CPU) hierarchy. This architecture effectively virtualizes device memory, providing the model with much larger cache capacity while significantly conserving expensive GPU resources.
    \item We introduce a suite of cohesive technical designs to ensure efficient execution, featuring a \textit{dual-granularity storage abstraction} (Page-Chunk) to balance I/O efficiency, an \textit{asynchronous latency-hiding pipeline} to effectively overlap data movement with computation, and a \textit{locality-driven management policy} to enable non-blocking, zero-copy metadata-driven cache maintenance.
\squishend

\section{Preliminaries}
\label{sec:preliminary}

\begin{figure}[!t]
    \centering
    \includegraphics[width=0.95\linewidth]{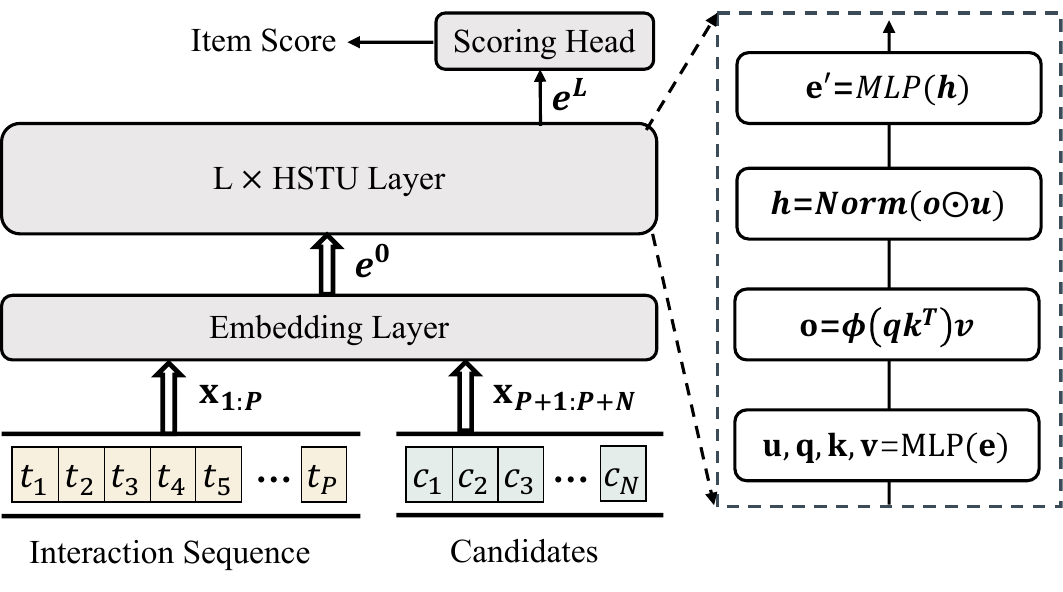} 
    \caption{The architecture of HSTU model for generative recommendation~\cite{zhai2024actions}.}
    \label{fig:model_arch}
\end{figure}

\stitle{Generative Recommendation Model (GRM).}
Generative recommendation models treat the ranking task as a sequence transduction problem. 
Given a user interaction sequence $\mathbf{H}_u$ comprising $P$ historical tokens and a candidate set 
$\mathcal{C} = \{c_1, c_2, \dots, c_N\}$, 
the model aims to estimate the relevance score $\gamma_i$ for each candidate conditioned on the historical context. 
As illustrated in Figure~\ref{fig:model_arch}, the input sequence $\mathbf{x}$ is constructed by concatenating the interaction sequence and candidates:
\begin{equation}
    \mathbf{x} = [\mathbf{x}_{1:P}; \mathbf{x}_{P+1:P+N}]
\end{equation}
where $\mathbf{x}_{1:P} = \mathbf{H}_u$ and $\mathbf{x}_{P+1:P+N} = \mathcal{C}$, resulting in a total sequence length of $T = P + N$.

The architecture consists of $L$ stacked blocks.
An initial embedding layer projects each token $x_t$ into a $d$-dimensional vector $\mathbf{e}_t^{(0)}$. 
Each block $\Psi^{(l)}$ ($1 \le l \le L$) performs a non-linear transformation. Following the state-of-the-art HSTU architecture~\cite{zhai2024actions}, 
each HSTU block functions similarly to a standard Transformer block, centered around an attention-based computation.
The internal computation of layer $l$ is formulated as:
\begin{equation}
    \mathbf{u}^{(l)}, \mathbf{q}^{(l)}, \mathbf{k}^{(l)}, \mathbf{v}^{(l)} = \text{Split}(\phi_1(\text{Linear}(\mathbf{e}_{1:T}^{(l-1)})))
\end{equation}
\begin{equation}
    \mathbf{o}^{(l)} = \phi_2(\text{Attention}(\mathbf{q}^{(l)}, \mathbf{k}^{(l)}, \mathbf{v}^{(l)}))
\end{equation}
\begin{equation}
    \mathbf{e}_{1:T}^{(l)} = \text{MLP}(\text{Norm}(\mathbf{o}^{(l)} \odot \mathbf{u}^{(l)}))
\end{equation}
where $\phi_1$ and $\phi_2$ denote the SiLU activation functions.
The hidden representations undergo $L$ such transformations to produce the final contextualized embeddings:
\begin{equation}
    \mathbf{e}_{1:T}^{(L)} = \Psi^{(L)} \circ \dots \circ \Psi^{(1)}(\mathbf{e}_{1:T}^{(0)})
\end{equation}

Focusing on the terminal position $T$, the representation $\mathbf{e}_T^{(L)}$ is fed into a \emph{scoring head} to obtain the ranking logits $\mathbf{y} = \mathbf{W}_{out} \mathbf{e}_T^{(L)}$.
For any candidate $c_i$ associated with its unique identifier token $\omega_i \in \mathcal{V}$, its relevance is directly determined by the corresponding logit value $y[\omega_i]$. The final ranking result is produced by sorting the candidates $\{c_1, \dots, c_N\}$ in descending order of their logit values.

\begin{figure*}[!t]
    \centering
    \includegraphics[width=0.8\linewidth]{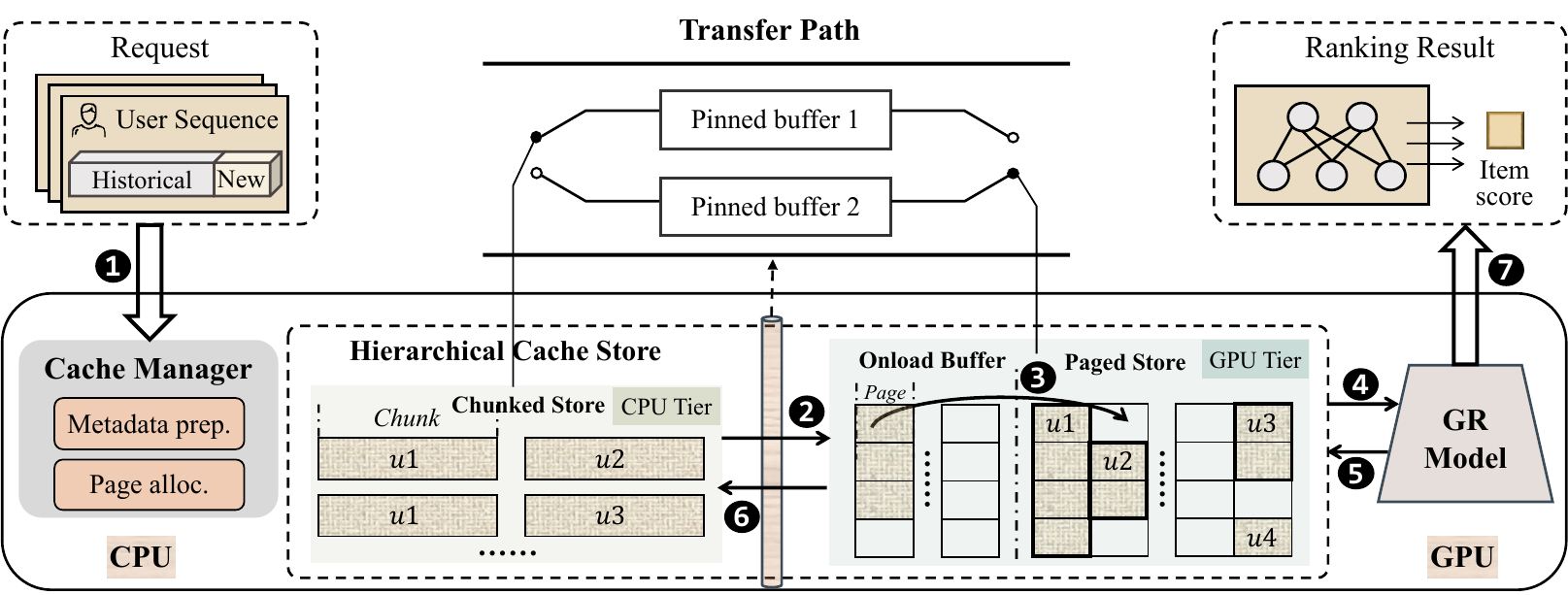} 
    \caption{The overall architecture and seven-step inference workflow of \method.}
    \label{fig:workflow}
\end{figure*}

\stitle{Serving GRMs with KV Cache.}
To serve GRMs efficiently, we decompose the user history into a cached historical prefix $\mathbf{H}_{pre}$ (length $P_{pre}$) and incremental tokens $\Delta \mathbf{H}$ (length $\Delta P$) that arrived since the last visit, such that $P = P_{pre} + \Delta P$.

When a user request arrives, the system retrieves the previously computed Key-Value tensors $(\mathbf{K}_{pre}^{(l)}, \mathbf{V}_{pre}^{(l)})$ for $\mathbf{H}_{pre}$ from the cache store. The incremental input $\mathbf{x}_{\Delta} = [\Delta \mathbf{H}; \mathcal{C}]$ is first projected into the initial embedding space $\mathbf{e}_{\Delta}^{(0)}$. For each subsequent layer $l$, the model only needs to compute the projections for the hidden representations $\mathbf{e}_{\Delta}^{(l-1)}$ derived from these new segments:
\begin{equation}
    \mathbf{u}_{\Delta}^{(l)}, \mathbf{q}_{\Delta}^{(l)}, \mathbf{k}_{\Delta}^{(l)}, \mathbf{v}_{\Delta}^{(l)} = \text{Split}(\phi_1(\text{Linear}(\mathbf{e}_{\Delta}^{(l-1)})))
\end{equation}
The complete context for the attention-like mechanism at layer $l$ is then assembled by concatenating the retrieved cache with the newly generated states:
\begin{equation}
    \mathbf{K}_{total}^{(l)} = [\mathbf{K}_{pre}^{(l)}; \mathbf{k}_{\Delta}^{(l)}], \quad \mathbf{V}_{total}^{(l)} = [\mathbf{V}_{pre}^{(l)}; \mathbf{v}_{\Delta}^{(l)}]
\end{equation}

This \textit{user-side caching} effectively decouples the inference latency from the total history length and reduces the effective attention complexity from $O(T^2)$ to $O((T-P_{pre}) \cdot T)$ per inference step. Finally, the updated KV states for the newly processed tokens are stored back to the cache to facilitate future reuse.

However, realizing this reuse in GR presents unique challenges. 
First, the combination of a massive user base and long interaction histories creates an immense storage demand. Unlike LLM serving where common prefixes can be optimized via radix trees~\cite{zheng2024sglang}, GR requires \textit{individual persistence} for millions of unique user sequences. This lack of sharing opportunities leads to a storage explosion that far outstrips physical GPU memory capacity. 
Second, the \textit{cross-request} nature of recommendation demands long-term durability across intermittent visits, rendering standard session-based, volatile caching insufficient. 
Finally, the \textit{high I/O-to-computation ratio} of relatively shallow GR models, coupled with frequent state swapping for short-lived requests, creates a severe I/O bottleneck that can easily stall the inference pipeline.

\section{MTServe Overview}
\label{sec:overview}

We design \method to tackle the GPU memory bottleneck in generative recommendation, which stems from the dual challenges of a massive user base and long interaction histories. 
Functioning as a plug-and-play cache management module, \method integrates seamlessly with attention-based generative models to enable efficient user-side KV cache reuse. 
Figure~\ref{fig:workflow} illustrates the overall architecture and the seven-step inference workflow.

\subsection{System Components}
\method comprises four primary components that collaborate to execute high-throughput inference by efficiently managing and utilizing the hierarchical KV cache.

\squishlist
    \item \textbf{Generative Recommendation (GR) Model.} The core inference engine (e.g., HSTU~\cite{zhai2024actions}). It interacts with the logical cache interface provided by the KV cache manager, remaining agnostic to the underlying physical storage and tiering logic.

    \item \textbf{KVCacheManager.} The central controller residing on the host. 
    It handles metadata management, including sequence length tracking, LRU replacement, and page table maintenance. 
    The page table acts as a logical-to-physical mapping structure where each entry tracks the physical location of a fixed-length segment (e.g., a 32-token page) of a user's KV cache, serving as the atomic unit for GPU memory management.

    \item \textbf{Hierarchical Cache Store.} 
    A dual-tier storage backend (as shown in the center of Figure~\ref{fig:workflow}). 
    It consists of the \textit{CPU Tier} (Chunked Store) as a scalable backup store and the \textit{GPU Tier} (Paged Store) as the primary cache for low-latency access. 
    Both tiers are managed via a paging/chunking mechanism to eliminate internal fragmentation.

    \item \textbf{Data Transfer Path.} A dedicated pipeline for moving KV blocks between tiers. 
    It utilizes \textit{Pinned Memory Buffers} on the CPU to enable high-speed DMA and \textit{Onload/Offload Buffers} on the GPU for data staging. 
    This path supports asynchronous, double-buffered transfers to overlap I/O with computation.
\squishend

\subsection{Inference Workflow}
\label{subsec:workflow}

The lifecycle of an inference request in \method follows a streamlined seven-step workflow, as depicted by the circled numbers in Figure~\ref{fig:workflow}. 
In our serving scenario, each incoming request encapsulates a unique user identifier, a sequence of incremental interaction tokens newly arrived since the user's last visit, and a set of candidate items to be ranked.




\squishlistcircled
    \item \textbf{Metadata Preparation.} Upon receiving a batch of requests, the \textit{KVCacheManager} prepares execution metadata by retrieving cached sequence lengths, resolving page mappings, and allocating new GPU pages. If device memory is exhausted, the LRU policy identifies victim pages for zero-copy eviction.

    \item \textbf{Data Onload.} For requests requiring historical data from the host (Cache Miss), the system asynchronously transfers relevant KV blocks from the \textit{CPU Tier} into a pinned memory buffer, then via DMA to the GPU's \textit{Onload Buffer}.

    \item \textbf{Cache Scattering.} Once the data resides in the Onload Buffer, the system executes an intra-GPU redistribution.
    A dedicated scatter kernel redistributes the contiguous data from the Onload Buffer into pre-allocated discrete pages within the \textit{GPU Tier}.

    \item \textbf{Model Inference.} With the cache populated, the GR Model executes the forward pass. 
    It utilizes the page table metadatato retrieve historical KV pairs from the Device Cache Store and combine them with current inputs for self-attention.

    \item \textbf{Cache Update.} The new KV states generated during the inference step are appended directly to the allocated pages in the Device Cache Store, updating the user's history in real time.

    \item \textbf{Cache Offload.} 
    A background process manages the write-back of new KV states. When a user's accumulated new tokens reach a \textit{host chunk size}, the corresponding GPU pages are gathered into a transiently allocated \textit{Offload Buffer}. This buffer is then asynchronously transferred to the  \textit{CPU Tier}, ensuring the CPU backend maintains a persistent record of the user history.

    \item \textbf{Ranking Result.} Finally, the hidden representations are passed to the scoring head to calculate relevance scores and produce the final ranking.
\squishendcircled

\section{Key Designs}
\label{sec:design}
This section details the system optimizations of \method, focusing on the synergy between its hierarchical storage abstraction, asynchronous data pipeline, and non-blocking cache management policy. Together, these designs maximize hardware utilization and ensure low-latency serving even under extreme memory pressure.

\subsection{Paged Hierarchical Storage}
\label{subsec:storage}

\textbf{Hybrid Storage Granularity.} 
Traditional contiguous memory allocation~\cite{nvidia2023fastertransformer} incurs severe fragmentation when handling variable-length user histories in recommendation systems. Specifically, static pre-allocation leads to massive \textit{internal fragmentation} as reserved slots for long-tail users remain underutilized, while dynamic resizing causes \textit{external fragmentation} that is computationally expensive to compact during high-QPS serving. Inspired by virtual memory systems and PagedAttention~\cite{kwon2023efficient}, \method implements a \emph{dual-granularity storage strategy} (Page-Chunk) tailored to the distinct hardware characteristics of the GPU and CPU tiers, as illustrated in the center of Figure~\ref{fig:workflow}.

\squishlist
    \item \textbf{GPU Tier (Fine-grained Paging):} On the GPU, where capacity is the primary bottleneck, we adopt a fine-grained page management strategy (e.g., 32 tokens per page). This granularity reduces internal fragmentation, allowing the cache to expand dynamically with the growing sequence length. The GPU-resident \textit{Paged Store} is defined as a pre-allocated tensor of shape $[L, N_{\text{pages}}, 2, S_{\text{page}}, H, D]$, where $L$ denotes layers, $N_{\text{pages}}$ total capacity, $S_{\text{page}}$ page size, and $H, D$ the attention head dimensions. By treating GPU memory as a pool of discrete pages, any free slot can be allocated to any user, effectively eliminating external fragmentation and improving HBM utilization.
    
    \item \textbf{CPU Tier (Coarse-grained Chunking):} Conversely, on the Host RAM, where capacity is abundant but bandwidth is constrained by the PCIe interface, 
    we manage data in coarser \emph{Chunks} (e.g., 1024 tokens). 
    Since PCIe throughput is highly sensitive to transaction overhead, transferring a single large chunk is significantly more efficient than moving multiple small, discontinuous pages. 
    This design amortizes the startup cost of DMA transactions,
    ensuring that the I/O subsystem can saturate the physical link during bulk data movement.
\squishend


\textbf{Hierarchical Indexing Mechanism.} To bridge these storage granularities, \method maintains a robust indexing architecture that acts as a logical-to-physical mapping structure. Each user is mapped to a list of non-contiguous \emph{Logical Page IDs} on the GPU and \emph{Logical Chunk IDs} on the CPU, allowing the sequence to appear contiguous to the attention mechanism while being physically scattered. A global page table translates these logical IDs into physical addresses—resolving to tensor indices for the GPU Tier and memory pointers for the CPU Tier—serving as the atomic unit for all memory operations. This mapping logic is replicated across layers to ensure \textit{layer-wise independence}, enabling the cascading execution pattern described in Section~\ref{subsec:pipeline} where different layers can simultaneously perform data loading, computation, and offloading.

\textbf{Functional Memory Regions.} 
As illustrated within the cyan box in Figure~\ref{fig:workflow}, the GPU memory is logically partitioned into two functional regions.
\squishlist
    \item \textbf{Device Cache Store (Primary Cache):} Represented as the \textit{Paged Store} in Figure~\ref{fig:workflow}, this is the main storage area in the GPU Tier holding persistent KV data for active users. It utilizes the discrete paged layout and is dynamically managed by the \textit{KVCacheManager} to ensure that the most frequently accessed user states remain on-device for low-latency retrieval. 
    \item \textbf{Onload Buffer:} A pre-allocated contiguous region serving a dual purpose. 
    Primarily, it acts as an active inference area where the model can directly access fresh data (e.g., newly loaded history from \textit{Host Cache Store}) without waiting for scattering. 
    Secondly, it serves as a \emph{staging ground} for data layout transformation, from which contiguous data blocks are eventually scattered into the discrete pages of the \textit{Device Cache Store}.
\squishend

\subsection{Efficient Data Movement}
\label{subsec:movement}

Efficiently moving KV blocks between host and device is critical for maintaining high system throughput, especially given the high I/O-to-computation ratio of recommendation models. We optimize this process by employing dedicated scatter/gather kernels for GPU memory layout transformation, reinforced by a double-buffered pipeline designed to saturate the physical PCIe bus bandwidth.

\stitle{Pipelined Onload via Double Buffering.} 
As illustrated in Figure~\ref{fig:workflow}, the onload process (Step \ding{173}) is internally structured into a two-stage pipeline. 
While represented as a single logical step in the workflow, 
it comprises two distinct physical sub-stages: 
(1) \textit{Host-side Memory Preparation}, where KV blocks are gathered from the chunked host store into a pinned memory buffer; 
and 
(2) \textit{PCIe Transfer} via DMA to the GPU's Onload Buffer. 
This is followed by the separate \textit{Cache Scattering} operation (Step \ding{174}), which redistributes the contiguous data from the Onload Buffer into discrete pages within the Device Cache Store.

To orchestrate this internal pipeline efficiently, \method employs a fine-grained \emph{double-buffering strategy} using a pair of pinned memory buffers (\textit{Pinned Buffer 1 \& 2} in Figure~\ref{fig:workflow}). Crucially, we align the size of each pinned buffer with the host storage \textit{Chunk} granularity. This alignment simplifies the host-side preparation into a high-speed sequential block copy, avoiding the heavy CPU overhead of gathering fragmented pages from the storage pool. Consequently, the system operates in a \textit{ping-pong} fashion: while the CPU is sequentially populating Buffer 1 with the next required KV chunk (sub-stage 1), the DMA engine simultaneously transfers the previously prepared Buffer 2 to the GPU (sub-stage 2). By overlapping host-side orchestration with hardware I/O, \method ensures that the DMA engine remains saturated and the transfer latency is largely hidden. Finally, a custom-optimized \textit{scatter kernel} (Step \ding{174}) distributes the contiguous data from the Onload Buffer into discrete physical pages within the GPU Tier.

\stitle{Transient Offload Pipeline.} 
To ensure long-term durability and enable non-blocking cache reclamation, \method employs a proactive offloading strategy rather than a reactive one. As shown in Step \ding{177} of Figure~\ref{fig:workflow}, the offload process operates as a background task to persist new KV states to the CPU Tier before they are targeted for eviction. By maintaining a consistent and up-to-date backup on the host, the system can reclaim GPU pages via simple metadata updates, avoiding the latency spikes associated with forced synchronization during memory pressure. 
The pipeline is triggered when a user's accumulated new tokens reach the pre-defined \textit{host chunk size}, initiating the following sequence: (1) A \textit{gather kernel} collects dispersed pages from the GPU Tier into a \textit{transiently allocated} GPU Offload Buffer; (2) The data is transferred via DMA to one of the host \textit{Pinned Buffers}; and (3) A \textit{multi-threaded CPU worker pool} copies the data from the Pinned Buffer to the final \textit{Chunked Host Cache}. Similar to the onload path, the use of dual Pinned Buffers allows the DMA transfer and the CPU-side copy to execute in parallel. This multi-threaded orchestration is crucial to match the high throughput of the DMA stream, ensuring that the proactive persistence remains strictly non-blocking and does not become a bottleneck for the primary inference stream.


\stitle{Offload Capacity Control.} 
To prevent GPU memory exhaustion from transient buffer accumulation, \method enforces a strict quota on the total volume of pending offload tasks. 
The system tracks the aggregate tokens currently queued in GPU transient buffers; 
if a new offload task would cause this total to exceed the pre-configured limit, the task is rejected via an admission control mechanism. 
This acts as a critical backpressure valve, ensuring that background maintenance tasks do not destabilize the primary inference service by consuming excessive VRAM, thereby guaranteeing system reliability even under extreme heavy traffic bursts.


\begin{figure}[!t]
    \centering
    \small
    \includegraphics[width=0.95\linewidth]{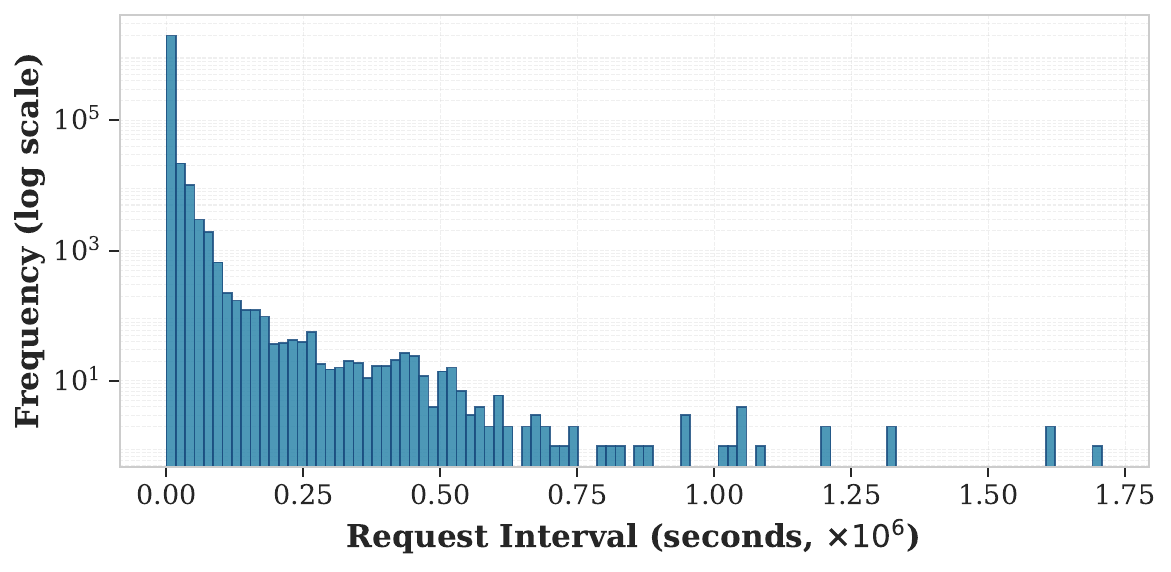} 
    \caption{
        Distribution of arrival intervals between consecutive requests from the same user. The heavy-tailed distribution indicates that most users return within a short period, exhibiting strong temporal locality.
    }
    \label{fig:interval_dist}
\end{figure}

\subsection{Cache Management}
\label{subsec:management}

Effective cache management hinges on exploiting the temporal patterns inherent in user traffic. 
As shown in Figure~\ref{fig:interval_dist}, the arrival intervals of consecutive requests from the same user exhibit a heavy-tailed distribution, 
where the majority of requests occur within a short temporal window. 
This strong temporal locality motivates the adoption of a Least-Recently-Used (LRU) policy, which prioritizes retaining recently active users in the limited GPU memory to maximize the cache hit ratio.


\stitle{Allocation and Eviction Policy.} 
To implement the LRU policy efficiently at scale, \method maintains a global doubly-linked list augmented with an auxiliary hash map to ensure $O(1)$ access and update complexity. This constant-time performance is vital for maintaining low latency as the number of managed user sessions grows into the millions. 
Page allocation is tightly coupled with this policy: when the system requires pages for a new request or incremental token generation, it first attempts to draw from the free pool. If the pool is exhausted, the system reclaims pages from the tail of the LRU list, targeting the least recently active users. Crucially, this reclamation is designed as a \emph{zero-copy metadata operation}. Since the Host Cache Store (CPU Tier) serves as the ultimate backup store and data persistence is handled asynchronously by the background offload process, victim pages on the GPU can be simply marked as free in the page table without triggering any blocking device-to-host transfer. This design ensures that cache maintenance tasks impose negligible latency overhead on the critical inference path.

\stitle{User Locking Protocol.} 
To ensure memory safety and data consistency during concurrent operations, we enforce a strict \textit{User Locking} protocol. Users currently undergoing background offload (Step \ding{177} in Figure~\ref{fig:workflow}), where their pages are being read by the DMA engine, are marked in a \textit{locked} registry. The LRU policy explicitly \emph{skips} these locked users during victim selection to prevent race conditions where physical pages might be prematurely reclaimed or overwritten while still being accessed by the transfer pipeline. This coordination guarantees that the memory hierarchy remains consistent even under high-concurrency workloads.

\subsection{Asynchronous Pipelining}
\label{subsec:pipeline}

We orchestrate the aforementioned components using a multi-stream execution model to maximize hardware utilization. 
The detailed end-to-end inference workflow and the corresponding pseudocode Algorithm~\ref{alg:hstu_inference_kvcache} are provided in Appendix~\ref{sec:pseudocode}.

\textbf{Four-Stream Concurrency.} 
\method utilizes four independent, non-blocking CUDA streams to overlap disparate workloads, effectively functioning as a producer-consumer system: 
(1) \emph{Compute Stream} for core model execution, including the Transformer-based forward pass (\emph{Step 8}); 
(2) \emph{Onload Stream} for host-to-device DMA transfers to retrieve historical KV chunks (initiated in \emph{Step 2}); 
(3) \emph{Scatter Stream} for intra-GPU data redistribution from the onload buffer to the paged store (initiated in \emph{Step 7}); 
and (4) \emph{Offload Stream} for background persistence by transferring new KV states back to the host (initiated in \emph{Step 9}). 
Dependencies between these streams are managed via fine-grained CUDA events to ensure correct execution order without global stalls. 
This multi-stream design allows the system to preprocess metadata and prepare embeddings (\emph{Steps 3--6}) in parallel with the heavy KV data movement occurring in the Onload Stream.

\textbf{Fine-Grained Layer-wise Synchronization.} 
A critical optimization in our design is the \textit{implicit layer-wise synchronization} (Step 8.3). 
Instead of a coarse-grained barrier that waits for the entire batch's data to be loaded, we implement a staggered execution mechanism using \texttt{KVOnloadHandle}. 
Before executing the attention mechanism for Layer $l$, the Compute Stream waits specifically for the completion event of Layer $l$'s scatter operation. 
This allows the computation of shallower layers to proceed immediately once their specific data is ready, even if the transfer for deeper layers is still in progress. 
By deeply overlapping the memory-bound data movement with the compute-bound attention mechanism, \method significantly reduces the end-to-end inference latency and mitigates the impact of the high I/O-to-computation ratio.

\section{Experimental Evaluation}
\label{sec:evaluation}

In this section, we conduct a comprehensive evaluation of \method to validate its effectiveness in generative recommendation scenarios. 
Specifically, we aim to answer the following three research questions:
\squishlist
    \item \textbf{RQ1 (Latency Reduction):} How effectively does \method reduce end-to-end inference latency compared to standard baselines, and how does its performance scale as the batch size increases?
    \item \textbf{RQ2 (Cache Efficiency):} How effectively does the hierarchical caching strategy maintain a high token hit ratio under constrained GPU memory budgets, and how does this translate to computational savings?
    \item \textbf{RQ3 (System Overhead):} Is the overhead introduced by complex metadata management and tiered data movement negligible compared to the massive computational gains?
\squishend

\begin{table*}[!t]
\centering
\caption{End-to-end latency (ms), speedup, and hit ratio comparison across batch sizes (BS) of 1, 4, and 8. \method consistently outperforms baselines on both the public KuaiRand-1K and the production MT dataset.}
\label{tab:overall_performance}
\resizebox{\linewidth}{!}{
\begin{tabular}{ll|ccc|ccc|ccc}
\toprule
\textbf{Dataset} & \textbf{Metric} & \multicolumn{3}{c|}{\textbf{BS=1}} & \multicolumn{3}{c|}{\textbf{BS=4}} & \multicolumn{3}{c}{\textbf{BS=8}} \\
\cmidrule(lr){3-5} \cmidrule(lr){6-8} \cmidrule(lr){9-11}
& \textbf{Method} & \textbf{Recomp.} & \textbf{GPU-Only} & \textbf{\method} & \textbf{Recomp.} & \textbf{GPU-Only} & \textbf{\method} & \textbf{Recomp.} & \textbf{GPU-Only} & \textbf{\method} \\
\midrule
\multirow{4}{*}{\textbf{KuaiRand-1K}} 
& Latency (ms) & 21.2 & 17.4 & \textbf{14.4} & 72.8 & 42.1 & \textbf{33.5} & 143.6 & 72.9 & \textbf{47.3} \\
& Speedup (vs. RE) & 1.00$\times$ & 1.22$\times$ & \textbf{1.47$\times$} & 1.00$\times$ & 1.73$\times$ & \textbf{2.17$\times$} & 1.00$\times$ & 1.97$\times$ & \textbf{3.04$\times$} \\
& GPU Hit Ratio & - & 63.52\% & 63.52\% & - & 63.92\% & 63.92\% & - & 64.36\% & 64.36\% \\
& Total Hit Ratio & - & 63.52\% & \textbf{94.66\%} & - & 63.92\% & \textbf{98.57\%} & - & 64.36\% & \textbf{98.59\%} \\
\midrule
\multirow{4}{*}{\textbf{MT}} 
& Latency (ms) & 14.1 & 12.8 & \textbf{11.7} & 43.6 & 28.9 & \textbf{17.3} & 82.4 & 48.6 & \textbf{26.6} \\
& Speedup (vs. RE) & 1.00$\times$ & 1.10$\times$ & \textbf{1.21$\times$} & 1.00$\times$ & 1.51$\times$ & \textbf{2.52$\times$} & 1.00$\times$ & 1.70$\times$ & \textbf{3.10$\times$} \\
& GPU Hit Ratio & - & 60.26\% & 60.26\% & - & 60.69\% & 60.69\% & - & 61.73\% & 61.73\% \\
& Total Hit Ratio & - & 60.26\% & \textbf{96.73\%} & - & 60.69\% & \textbf{97.44\%} & - & 61.73\% & \textbf{98.71\%} \\
\bottomrule
\end{tabular}
}
\end{table*}

\begin{table*}[!t]
\centering
\caption{Latency decomposition (ms) of the inference pipeline on KuaiRand-1K. The step labels correspond to the stages defined in \textbf{Algorithm~\ref{alg:hstu_inference_kvcache}}. \method significantly reduces the execution time of core inference (Step 8).}
\label{tab:latency_decomposition}
\resizebox{\linewidth}{!}{
\begin{tabular}{l|ccc|ccc|ccc}
\toprule
\textbf{Batch Size} & \multicolumn{3}{c|}{\textbf{BS=1}} & \multicolumn{3}{c|}{\textbf{BS=4}} & \multicolumn{3}{c}{\textbf{BS=8}} \\
\cmidrule(lr){2-4} \cmidrule(lr){5-7} \cmidrule(lr){8-10}
\textbf{Method} & \textbf{Recomp.} & \textbf{GPU-Only} & \textbf{\method} & \textbf{Recomp.} & \textbf{GPU-Only} & \textbf{\method} & \textbf{Recomp.} & \textbf{GPU-Only} & \textbf{\method} \\
\midrule
Step 1-2. Prepare Metadata & - & 0.1 & 0.1 & - & 0.1 & 0.1 & - & 0.1 & 0.1 \\
Step 3. Strip Tokens & - & 0.4 & 0.5 & - & 0.4 & 0.6 & - & 0.5 & 0.6 \\
Step 4. Embedding & 2.2 & 1.2 & 1.5 & 5.0 & 1.3 & 1.5 & 9.7 & 1.5 & 1.5 \\
Step 5. Data Layout & 1.9 & 1.6 & 2.0 & 3.4 & 2.4 & 2.9 & 5.7 & 3.6 & 3.6 \\
Step 6. Await Metadata & - & 0.01 & 0.1 & - & 0.1 & 0.1 & - & 0.1 & 0.1 \\
Step 7. Update Metadata & - & 0.05 & 0.2 & - & 0.05 & 0.4 & - & 0.05 & 0.7 \\
Step 8. HSTU Inference & 16.7 & 13.5 & \textbf{9.4} & 63.9 & 37.1 & \textbf{27.1} & 127.4 & 66.3 & \textbf{39.8} \\
Step 9. Offload KV & - & - & 0.03 & - & - & 0.03 & - & - & 0.03 \\
Step 10. Postprocess & 0.5 & 0.5 & 0.5 & 0.6 & 0.6 & 0.6 & 0.8 & 0.7 & 0.7 \\
\bottomrule
\end{tabular}
}
\end{table*}

\subsection{Experimental Settings}

\stitle{Model Architecture.} 
We employ the Hierarchical Sequential Transduction Unit (HSTU)~\cite{zhai2024actions} as our backbone model. HSTU is a state-of-the-art generative recommendation architecture designed to capture long-term user interests with high efficiency. While our evaluation focuses on HSTU, the proposed \method system is generic and compatible with other attention-based generative recommenders such as SASRec~\cite{kang2018self} and BERT4Rec~\cite{sun2019bert4rec}. The specific model configuration used in our experiments consists of 8 Transformer-based layers, 4 attention heads, and a hidden dimension of 128 per head, totaling a model size and complexity for large-scale deployment.

\stitle{Cache Configuration.} 
By default, \method is configured with a page size of 32 tokens, a primary cache of 40,960 pages, and an offload chunk size of 1024 tokens. 

\stitle{Baselines.} 
To rigorously evaluate the performance benefits, we compare \method against two representative serving paradigms.

\squishlist
\item \textbf{RE (Recompute):} This paradigm, exemplified by standard implementations of \textit{HSTU}~\cite{zhai2024actions} and \textit{OneRec}~\cite{deng2025onerec}, re-encodes the entire user interaction history for every inference request. It serves as the performance lower bound, representing systems without any cross-request KV caching optimization.
\item \textbf{GPU-Only Cache:} A standard industry practice that stores KV states exclusively in high-bandwidth GPU memory, 
like BAT~\cite{sun2026modelsystem}, and OneTrans~\cite{zhang2025onetrans}. 
Due to the limited VRAM capacity, it employs a strict LRU eviction policy. Once the GPU memory is full, the least recently used user sessions are evicted and permanently lost, 
forcing an expensive re-computation upon their next arrival.
\squishend

\stitle{Evaluation Metrics.}
We evaluate the system using \emph{Average Latency} (ms) and \emph{Token Hit Ratio} (\%), where the latter represents the percentage of historical tokens retrieved from the hierarchical cache relative to the total sequence length.

\stitle{Datasets and Preprocessing.} 
We utilize two datasets to evaluate \method under different scenarios. 
(1) KuaiRand-1K\footnote{KuaiRand Dataset: \url{https://kuairand.com/}}: A public sequential recommendation benchmark from the video-sharing platform Kuaishou. To simulate a realistic streaming workload, we organize user behaviors by timestamp and segment them into individual requests (e.g., 60-second intervals), mimicking the temporal arrival patterns of online services. 
(2) MT Dataset: A production dataset featuring approximately 3,000 users with exceptionally long interaction histories. We specifically filter for long-sequence interactions to assess the system's robustness under sustained memory pressure and its ability to maximize history reuse. Detailed statistics for both datasets are provided in Appendix~\ref{sec:dataset}.

\stitle{Hardware and Implementation.}
All experiments are conducted on a high-performance server node equipped with an NVIDIA A100 GPU (80GB HBM2). The host side features 184 CPU cores and 928 GB of DDR4 RAM, serving as the backup store for our hierarchical cache. \method is implemented using PyTorch, with critical performance components such as scatter/gather kernels and metadata management implemented as custom CUDA and C++ extensions to minimize software-side overhead.

\subsection{Overall Performance (RQ1 \& RQ2)}
We evaluate the end-to-end performance and latency decomposition of \method across different workloads, including a smaller model variant (detailed in Appendix~\ref{app:small_model}).

\textbf{RQ1: Latency Reduction.} 
As shown in Table~\ref{tab:overall_performance}, \method consistently achieves the lowest latency across all batch sizes and datasets, demonstrating superior scalability compared to both RE and GPU-Only baselines. 
On KuaiRand-1K, \method provides a \textbf{1.47$\times$} speedup at BS=1, which scales to \textbf{3.04$\times$} at BS=8 (47.3 ms vs. 143.6 ms). 
The performance gain is even more pronounced on the production MT dataset, where \method achieves a \textbf{3.10$\times$} speedup at BS=8 (26.6 ms vs. 82.4 ms). 
The higher recompute latency on KuaiRand-1K (143.6 ms vs. 82.4 ms) is due to its much longer sequences (Table~\ref{tab:dataset_statistics}). The 20,000-token sequences in KuaiRand-1K create a significant bottleneck that slows down the entire batch.

The primary source of this gain is further revealed in the latency decomposition (Table~\ref{tab:latency_decomposition}). 
The computational cost in \textit{Step 8 (HSTU Inference)} is drastically reduced; 
at BS=8, \method completes the core inference in \textbf{39.8 ms}, while the RE baseline requires \textbf{127.4 ms} due to the exhaustive re-encoding of the entire user history. 
This trend confirms that \method successfully decouples inference latency from the total history length, transforming the generative recommendation task from a compute-intensive encoding problem into an efficient, retrieval-based process.

\textbf{RQ2: Cache Efficiency and Hit Ratio.} 
The performance advantage of \method is rooted in its ability to maintain a near-perfect hit ratio by breaking the GPU memory wall. 
While the \textit{GPU-Only} baseline is capped at a GPU hit ratio of $\approx$\textbf{60\%--64\%} due to VRAM limits (Table~\ref{tab:overall_performance}), \method utilizes host RAM as a backup store to achieve a total hit ratio of \textbf{94.6\%--98.7\%}. 

Specifically, on the MT dataset, the host tier successfully supplements the \textbf{61.73\%} GPU hit ratio to reach a \textbf{98.71\%} total hit ratio at BS=8. 
This high hit ratio remains stable across batch sizes, demonstrating that our LRU-based policy and hierarchical storage successfully capture the temporal locality of user traffic. 
By shifting the workload from redundant re-computation to efficient KV retrieval, \method significantly reduces the execution time of the attention mechanism, achieving a \textbf{1.67$\times$} speedup in \textit{Step 8} compared to the GPU-Only baseline at BS=8 (39.8 ms vs. 66.3 ms).
These results validate \method's readiness for industrial-scale deployment, where the active working set of user states frequently exceeds physical device limits.

\begin{table}[!t]
\centering
\caption{Impact of Chunk Size on inference latency (ms).}
\label{tab:chunk_size_impact}
\small
\begin{tabular}{l|cccc}
\toprule
\textbf{Chunk Size} & \textbf{512} & \textbf{1024} & \textbf{2048} & \textbf{4096} \\
\midrule
Wait Time (I/O) & 34.64 & 23.11 & 22.22 & 19.09 \\
Comp Time (Kernel) & 12.66 & 13.35 & 14.79 & 17.74 \\
\midrule
\textbf{End2end Latency} & 57.6 & 47.3 & 49.2 & 49.9 \\
\bottomrule
\end{tabular}
\end{table}

\subsection{System Overhead Analysis (RQ3)}
We examine the fine-grained execution costs of each pipeline stage to assess the system overhead introduced by \method. As shown in the latency decomposition in Table~\ref{tab:latency_decomposition}, the system maintains high efficiency by keeping control-plane operations minimal.

\squishlist
    \item \textbf{Metadata and Control Overhead:} The combined CPU overhead for metadata preparation, page table resolution, and synchronization (\textit{Steps 1, 2, 6, and 7}) remains remarkably low, totaling less than \textbf{1 ms} even at BS=8. 
    This confirms that our chunk-based indexing and zero-copy eviction policy introduce negligible computational burden.
    
    \item \textbf{Asynchronous Offload Efficiency:} The initiation of the background offload process (\textit{Step 9}) consumes only \textbf{0.03 ms}. 
    As designed in Section~\ref{subsec:movement}, the actual KV transfer is handled by background threads, ensuring that the persistence of user states imposes zero blocking latency on the critical inference path.

    \item \textbf{I/O vs. Computation Trade-off:} \method trades expensive re-computation (127.4 ms) for significantly cheaper KV restoration (39.8 ms). This 39.8 ms represents the \textit{effective critical path} where incremental computation overlaps with I/O. The net saving of \textbf{87.6 ms} proves that the I/O penalty of moving data is far outweighed by the computational savings of attention reuse, confirming that our pipeline successfully masks the majority of transfer overhead.
\squishend

\subsection{Sensitivity and Micro-Benchmarks}
\label{subsec:micro_experiments}

\stitle{Impact of Offload Chunk Size.} 
We examine the impact of \textit{Offload Chunk Size} on the trade-off between data movement and computation. To quantify this, we instrument the core HSTU inference loop (\textit{Step 8}) to separately measure the synchronization overhead (\textit{Wait Time}) and the kernel execution time (\textit{Comp Time}). 

As shown in Table~\ref{tab:chunk_size_impact}, increasing the chunk size from 512 to 4096 tokens significantly reduces the \textit{Wait Time} (34.64 ms $\to$ 17.09 ms). This reduction is driven by two factors: first, a coarser granularity amortizes fixed PCIe transaction overhead; second, larger chunks reduce the total data volume transferred by allowing longer "tail" segments to remain on the GPU without being persisted to the host. 
However, this I/O saving comes at the cost of increased \textit{Comp Time} (12.66 ms $\to$ 17.74 ms), a phenomenon governed by the \emph{persistence threshold}. Since \method only backs up KV states to the CPU tier once they accumulate into a full chunk, any un-offloaded tail tokens are lost if the user is evicted from the GPU. Consequently, these missing segments must be re-encoded from scratch during the next visit to reconstruct the sequence context, thereby increasing the computational burden. We select 1024 as the default configuration,
as it captures the steepest drop in Wait Time while minimizing the extra computational burden associated with larger chunks.

\begin{figure}[!t]
    \centering
    \includegraphics[width=0.95\linewidth]{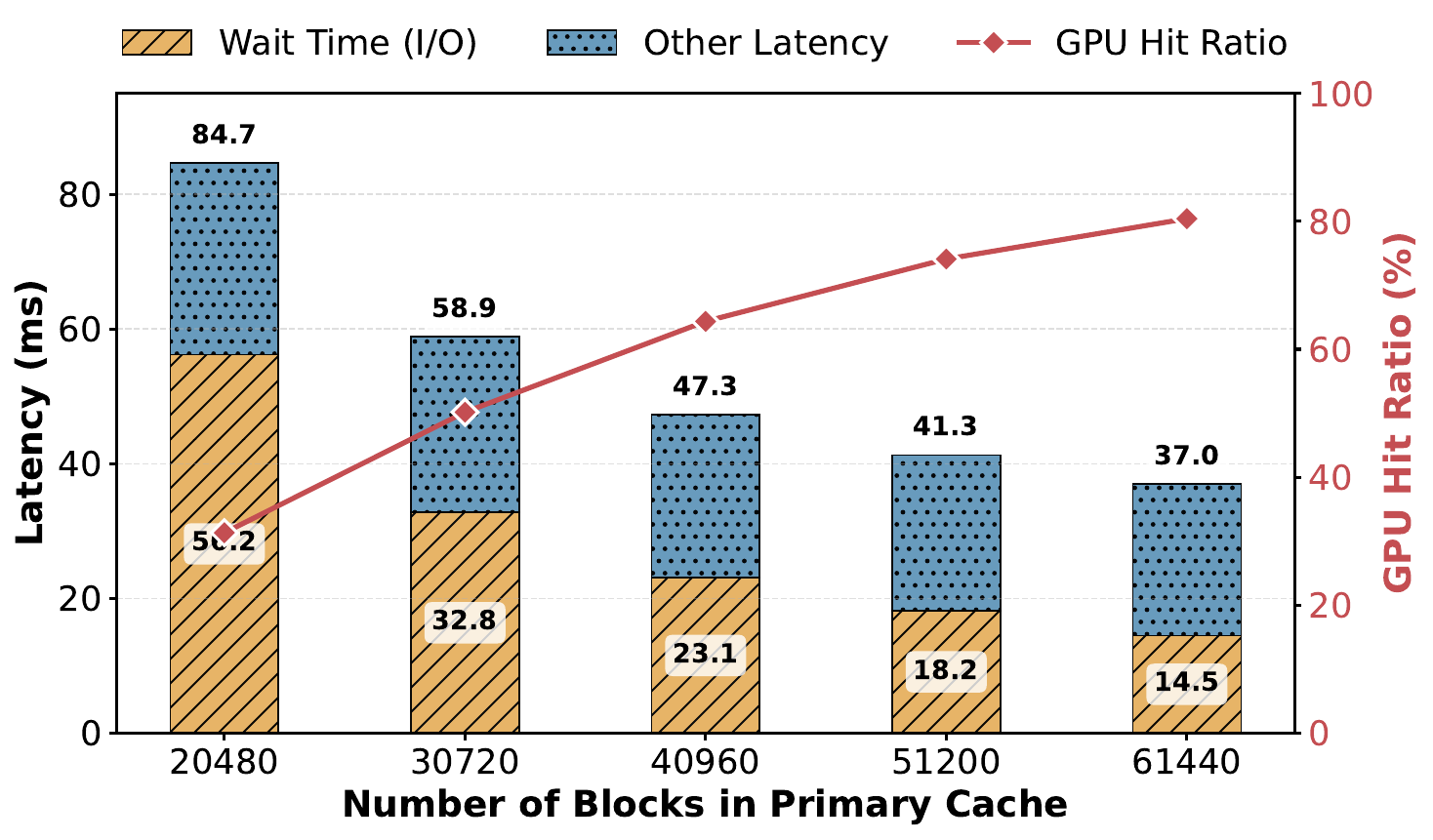} 
    \caption{Impact of GPU Cache Store capacity on on latency components and cache hit ratio.
    }
    \label{fig:capacity_impact}
\end{figure}

\stitle{Impact of GPU Cache Capacity.} 
We evaluate system robustness by adjusting the \textit{Primary Cache} capacity from 20,480 to 61,440 blocks. 
As shown in Figure~\ref{fig:capacity_impact}, increasing the capacity significantly improves the \textit{GPU Hit Ratio}, 
leading to a sharp decline in I/O wait time. However, this performance gain comes at the cost of a higher memory footprint; as detailed in Table~\ref{tab:capacity_sensitivity}, the VRAM consumption scales from 10 GB to 30 GB across these settings. 
Notably, even at the most constrained setting (20,480 blocks, 10 GB VRAM), \method still achieves an end-to-end latency of \textbf{84.7 ms}, representing a \textbf{1.69$\times$} speedup over the RE baseline (143.6 ms). 

\stitle{Memory Footprint Analysis} 
We profile the peak GPU memory usage during inference to identify the primary resource consumers. Our analysis reveals that the KV Cache is the dominant component, occupying over 59\% of the total 41.67 GB footprint. 
A detailed breakdown of the memory components is provided in Appendix~\ref{app:memory_profile}.

\section{Conclusion and Future Directions}
\label{sec:conclusion}

This paper presented \method, a hierarchical KV cache system that overcomes the GPU memory wall for reusing the KV caches in serving generative recommendation models. \method achieves high efficiency with extensive system optimizations including paged hierarchical storage, efficient data movement, and asynchronous pipelining. Experiment results show that \method can speed up model inference by up to a \textbf{3.1$\times$}.

Moving forward, we will explore three interesting directions: (i) applying compression techniques to the KV caches to improve the effective storage capacity and reduce data movement costs; (ii) deepening the storage hierarchy to include local NVMe SSDs to enlarge cache size and store cold users; (iii) enabling cross-machine transfer of the KV caches for scheduling and load balance.



\clearpage
\bibliographystyle{ACM-Reference-Format}
\bibliography{sample-base}

\clearpage
\appendix

\section{Pseudocode}
\label{sec:pseudocode}

Algorithm~\ref{alg:hstu_inference_kvcache} illustrates the inference pipeline of the generative recommendation model integrated with our hierarchical KV cache manager. 
Specifically, the pipeline initiates metadata preparation and KV onloading in the background (Lines 1--2) while concurrently processing input embeddings (Lines 3--5). Metadata tasks, such as page allocation and LRU-driven eviction, are offloaded to the CPU to avoid stalling the main inference thread. Synchronization is performed implicitly within the transformer layers (Line 8.3) to ensure data availability before attention computation, allowing the transfer of deeper KV states to overlap with the computation of initial layers. Finally, the system performs a non-blocking asynchronous offload of the updated cache (Line 9), ensuring state persistence without extending the request latency.

\begin{algorithm*}[b]
    \caption{Overall HSTU Inference Pipeline using Asynchronous Paged KV Cache Management.}
    \label{alg:hstu_inference_kvcache}
    
    \begin{algorithmic}[1]
        \REQUIRE $Input\text{ }Batch$: a batch of user and feature IDs; \\
                 $AsyncKVCacheManager$: a manager for offloading/onloading KVs, page allocation, and LRU eviction.
        \ENSURE $Output\text{ }Logits$: final ranking scores for all users in the batch.
        
        \STATE \textbf{1. Fetch Cache Lengths:} Retrieve users' total historical sequence length.
            \STATE \quad $TotalHistLengthes \gets \text{AsyncKVCacheManager.get\_total\_cache\_length}(Input Batch.users)$
            
        \STATE \textbf{2. Asynchronously Prepare Metadata and Onload KVs:} Initiate background page allocation, LRU ranking, and \textit{asynchronous} transfer of historical KVs from host to GPU onload buffer.
            \STATE \quad $MetadataFut, OnloadFut \gets \text{AsyncKVCacheManager.SubmitPrepareMetadataAndOnloadAsync}(\text{TotalHistLengthes}, \text{batch\_size})$
        
        \STATE \textbf{3. Preprocess Input Tokens:} Strip cached history tokens from inputs.
            \STATE \quad $StrippedBatch \gets \text{AsyncKVCacheManager.strip\_cached\_tokens}(Input Batch)$
            
        \STATE \textbf{4. Compute Embeddings:} Generate embeddings for non-cached inputs.
            \STATE \quad $Embeddings \gets \text{EmbeddingCollection}(StrippedBatch.features)$
        
        \STATE \textbf{5. Transform Data Layout:} Convert embeddings into JaggedArray for HSTU input.
            \STATE \quad $JaggedData \gets \text{HSTUBlock.preprocessor}(Embeddings, StrippedBatch)$
        
        \STATE \textbf{6. Await Metadata (Blocking):} Wait until metadata preparation (page allocation, LRU) is fully complete. \textit{(Note: Onload operation proceeds asynchronously in the background.)} 
            \STATE \quad $KVCacheMetadata \gets \text{AsyncKVCacheManager.AwaitMetadata}(MetadataFut)$
            
        \STATE \textbf{7. Update Metadata and Commit:} Update metadata to include lengths of the newly arrived batch sequence and commit the onloaded data from the buffer into the main paged cache.
            \STATE \quad $KVCacheMetadata.total\_history\_offsets \mathrel{+}= JaggedData.num\_offsets$
            \STATE \quad $KVCacheMetadata.total\_history\_lengths \mathrel{+}= JaggedData.num\_candidates$
            \STATE \quad $\text{AsyncKVCacheManager.CommitOnloadBufferToCache}(Input Batch.users)$
        
        \STATE \textbf{8. HSTU Core Computation Loop:} Execute Attention computation for each Transformer layer $l$.
        \FOR{each layer $l$ in HSTU model}
            \STATE \quad \textbf{8.1. Compute and Reshape QKV:} $Q, K, V \gets \text{ComputeQKV}(JaggedData.values)$
            \STATE \quad \textbf{8.2. Append New KVs to Paged Cache:} Store $K, V$ for the current input into the Paged KV Cache.
                \STATE \quad \quad 
                $ append\_kvcache(\text{K, V, }KVCacheMetadata\text{.position, }KVCacheMetadata\text{.offsets,}$
                \STATE \quad \quad \quad \quad \quad \quad \quad \quad \quad \quad $KVCacheMetadata\text{.indices/indptr/last\_page\_lens})$
            \STATE \quad \textbf{8.3. Implicit Sync for Historical KVs (Critical):} Implicitly wait for the historical KV data specific to layer $l$ to be fully loaded from the onload buffer into the paged cache. This synchronization is managed by the layer's \textit{KVOnloadHandle}.
            \STATE \quad \textbf{8.4. Compute Attention Scores:} $OutputValues \gets \text{HSTUAttention}(Q, K, V)$
            \STATE \quad \textbf{8.5. Fuse and Feed-Forward:} $JaggedData.values \gets \text{PostAttentionProcess}(OutputValues)$
        \ENDFOR
        
        \STATE \textbf{9. Asynchronously Offload KVs:} Background save the updated KV cache to host memory.
            \STATE \quad $\text{AsyncKVCacheManager.SubmitOffloadAsync}(KVCacheMetadata)$
            
        \STATE \textbf{10. Final Outputs:} Finalize outputs and compute final ranking logits.
            \STATE \quad $HSTUOut \gets \text{HSTUBlock.postprocessor}(JaggedData.values)$
            \STATE \quad $FinalLogits \gets \text{MLP}(HSTUOut)$
        
        \RETURN $Output Logits: (FinalLogits)$
    \end{algorithmic}
\end{algorithm*}

\section{Dataset Statistics}
\label{sec:dataset}

\begin{table}[h]
\centering
\caption{Statistics of the Evaluation Datasets.}
\label{tab:dataset_statistics}
\begin{tabular}{cccccc}
\toprule
\textbf{Dataset} & \textbf{\#Users} & \textbf{\#Reqs.} & \textbf{Min} & \textbf{Max} & \textbf{Avg} \\
\midrule
\makecell[c]{Kuai\\Rand-1k} & 1,000 & 1,181,699 & 1 & 20,000 & 6,375 \\
MT & 2,884 & 1,972,438 & 4,000 & 6,000 & 5,189 \\
\bottomrule
\end{tabular}
\end{table}

\label{app:dataset_stats}
This section provides the key statistics for the evaluation datasets. Table~\ref{tab:dataset_statistics} summarizes the user count, total requests, and the distribution of sequence lengths.
\newline

\begin{table*}[t]
\centering
\caption{Detailed latency (ms) and hit ratio for a 4-layer HSTU Model ($L=4$) on KuaiRand-1K.}
\label{tab:small_model_latency}
\small
\begin{tabular}{l|ccc|ccc|ccc}
\toprule
\textbf{Batch Size} & \multicolumn{3}{c|}{\textbf{BS=1}} & \multicolumn{3}{c|}{\textbf{BS=4}} & \multicolumn{3}{c}{\textbf{BS=8}} \\
\cmidrule(lr){2-4} \cmidrule(lr){5-7} \cmidrule(lr){8-10}
\textbf{Method} & \textbf{Recomp.} & \textbf{GPU-Only} & \textbf{\method} & \textbf{Recomp.} & \textbf{GPU-Only} & \textbf{\method} & \textbf{Recomp.} & \textbf{GPU-Only} & \textbf{\method} \\
\midrule
Step 1-2. Prepare Metadata & -- & 0.1 & 0.1 & -- & 0.1 & 0.1 & -- & 0.1 & 0.1 \\
Step 3. Strip Tokens & -- & 0.4 & 0.5 & -- & 0.4 & 0.5 & -- & 0.5 & 0.6 \\
Step 4. Embedding & 2.2 & 1.2 & 1.2 & 1.5 & 1.3 & 1.3 & 1.9 & 1.5 & 1.3 \\
Step 5. Data Layout & 1.9 & 1.6 & 1.6 & 3.4 & 2.4 & 2.5 & 5.7 & 3.6 & 3.3 \\
Step 6. Await Metadata & -- & 0.1 & 0.1 & -- & 0.1 & 0.1 & -- & 0.1 & 0.1 \\
Step 7. Update Metadata & -- & 0.05 & 0.1 & -- & 0.05 & 0.2 & -- & 0.05 & 0.4 \\
Step 8. HSTU Inference & 16.7 & 6.9 & \textbf{5.1} & 31.9 & 18.7 & \textbf{11.3} & 64.1 & 33.4 & \textbf{20.4} \\
Step 9. Offload KV & -- & -- & 0.02 & -- & -- & 0.03 & -- & -- & 0.03 \\
Step 10. Postprocess & 0.5 & 0.5 & 0.5 & 0.6 & 0.5 & 0.6 & 0.7 & 0.6 & 0.7 \\
\midrule
\textbf{Total Latency (ms)} & 21.2 & 10.8 & \textbf{9.2} & 37.4 & 23.7 & \textbf{16.7} & 72.4 & 39.9 & \textbf{27.0} \\
\textbf{Speedup vs. Recomp.} & 1.00$\times$ & 1.96$\times$ & \textbf{2.30$\times$} & 1.00$\times$ & 1.58$\times$ & \textbf{2.24$\times$} & 1.00$\times$ & 1.81$\times$ & \textbf{2.68$\times$} \\
\textbf{Speedup vs. GPU-Only} & -- & 1.00$\times$ & \textbf{1.17$\times$} & -- & 1.00$\times$ & \textbf{1.42$\times$} & -- & 1.00$\times$ & \textbf{1.48$\times$} \\
\midrule
\textbf{Total Hit Ratio} & -- & 63.52\% & \textbf{98.56\%} & -- & 63.92\% & \textbf{98.57\%} & -- & 64.34\% & \textbf{98.59\%} \\
\bottomrule
\end{tabular}
\end{table*}

\textbf{KuaiRand-1k.} This dataset exhibits a high variance in sequence lengths (ranging from 1 to 20,000 tokens), representing a natural distribution of user behaviors. It is used to evaluate the system's adaptability across different workloads.

\textbf{MT Dataset.} The MT dataset is a production trace from \textit{Meituan}. As shown in Table~\ref{tab:dataset_statistics}, we specifically filtered this dataset to focus on users with \textit{long-history} interactions (4,000--6,000 tokens). By eliminating short-sequence noise, the MT dataset provides a stable environment to assess the performance of \method under sustained memory pressure.

\section{Detailed Memory Footprint Analysis}
\label{app:memory_profile}
This section details the GPU memory allocation for \method during active inference ($B=8, L_{max}=40,008$) using \texttt{bfloat16} precision. The total peak footprint of \emph{41.67 GiB} (42,671 MiB), primarily driven by the persistent KV cache and the pre-allocated inference workbenches.

\subsection{Persistent KV Cache Storage ($\sim$24.89 GiB)}
This component represents the historical user state retained for cross-request reuse. It is pre-allocated as a unified \texttt{cache\_table} tensor with the shape $[L, (N_{p} + N_{o}), 2, S_{page}, H, D]$, where $L=8$, $S_{page}=32$, $H=4$, and $D=128$. The allocation comprises two functional regions:
\squishlist
    \item \textbf{Primary Cache ($N_p$):} A fixed pool of 40,960 pages reserved for active users.
    \item \textbf{Onload Buffer ($N_o$):} 
    The staging area capacity is determined by the maximum tokens per batch to ensure all incoming data can be accommodated: 
    $N_o = \lceil (B \times L_{max}) / S_{page} \rceil = \lceil 320,064 / 32 \rceil \approx 10,008$ pages.
\squishend
The total allocation for these 50,968 pages ($N_p + N_o$) occupies approximately \textbf{25,484 MiB} ($8 \times 50,968 \times 2 \times 32 \times 4 \times 128 \times 2$ bytes).

\subsection{Inference Workbench ($\sim$14.65 GiB)}
To eliminate dynamic memory management overhead, each of the 8 Transformer layers pre-allocates a dedicated "workbench" for intermediate activations. The size is governed by the peak token capacity $T_{max} = B \times L_{max} = 320,064$. Each layer reserves approximately \textbf{1,875 MiB}, broken down as follows:
\squishlist
    \item \textbf{UVQK Projection Buffer (1,250 MiB):} Stores the intermediate $U, V, Q, K$ projections. With a hidden dimension of 128 per head and 4 heads, this requires $T_{max} \times (128 \times 4 \times 4) \times 2 \approx 1,250$ MiB.
    \item \textbf{Output Buffer (625 MiB):} Stores the activation outputs of the attention and FFN blocks, requiring $T_{max} \times 1024 \times 2 \approx 625$ MiB.
\squishend
Across all 8 layers, this workbench occupies a total of \textbf{15,000 MiB}.

\subsection{Summary}
The remaining memory ($\sim$2,187 MiB) is occupied by static model weights and miscellaneous system buffers. This analysis confirms that the \emph{KV cache} and the \emph{layer-wise workbench} are the dominant memory consumers. By virtualizing the KV cache via host RAM, \method effectively caps the persistent state's GPU footprint, ensuring that sufficient memory remains available required by long-sequence generative models.

\begin{table}[!t]
\centering
\caption{End-to-end latency across different GPU Primary Cache capacities (BS=8). Reference \textit{Recompute} latency: \textbf{143.6 ms}.}
\label{tab:capacity_sensitivity}
\small
\setlength{\tabcolsep}{12pt}
\begin{tabular}{ccc}
\toprule
\textbf{\begin{tabular}[c]{@{}c@{}}Primary Cache\\ (\#Blocks)\end{tabular}} & 
\textbf{\begin{tabular}[c]{@{}c@{}}VRAM\\ (approx.)\end{tabular}} & 
\textbf{\begin{tabular}[c]{@{}c@{}}Total Latency\\ (ms)\end{tabular}} \\
\midrule
20,480 & 10 GB & \textbf{84.7} \\
30,720 & 15 GB & \textbf{58.9} \\
40,960 & 20 GB & \textbf{47.3} \\
51,200 & 25 GB & \textbf{41.3} \\
61,440 & 30 GB & \textbf{37.0} \\
\bottomrule
\end{tabular}
\end{table}

\section{Extended Experimental Results}
\label{app:extended_results}

\subsection{Performance on Smaller Model Workload}
\label{app:small_model}
To evaluate the adaptability of \method across different model scales, we conduct additional experiments on a smaller HSTU variant with 4 layers ($L=4$). This setup reduces computational intensity, allowing us to observe system efficiency when the computation-to-I/O ratio is lower. 

Table~\ref{tab:small_model_latency} presents the detailed latency breakdown and hit ratios. Similar to the observations in the 8-layer model, \method (\textit{\method}) maintains a significant performance lead over both \textit{GPU-Only} and \textit{Recompute} baselines. At BS=8, \method achieves a total latency of \textbf{27.0 ms}, representing a \textbf{2.68$\times$} speedup over recomputation. The consistency of the \textit{Total Hit Ratio} ($\sim$98.6\%) across model scales further validates that our tiered storage effectively virtualizes GPU memory regardless of the specific model depth.

\subsection{Sensitivity Analysis of GPU Cache Capacity}
\label{app:capacity_sensitivity}
Table~\ref{tab:capacity_sensitivity} provides a concise comparison of the system's performance under various GPU cache capacities. We observe a clear trade-off: while increasing the VRAM footprint to 30 GB minimizes latency to 37.0 ms, the system remains highly effective even at a 10 GB limit, highlighting the flexibility of our hierarchical storage design.





\section{Related Work}
\label{app:related_work}

\stitle{Generative Recommendation Models (GRMs).}
The paradigm of recommendation systems has significantly evolved in recent years. While deep learning-based methods~\cite{cheng2016wide, guo2017deepfm, wang2017deep, zhou2018deep, zhou2019deep, covington2016deep, perozzi2014deepwalk, wang2018billion} dominated for nearly a decade, they often face limitations in scalability and expressive power. The rise of large language models has ignited a shift towards generative recommendation~\cite{rajput2023recommender, zhai2024actions, liu2025generative}. This novel paradigm re-frames recommendation as a sequence transduction problem, leveraging the models' profound semantic understanding and open-ended generation capability to directly produce recommendations, explanations, or personalized content, thereby aiming for more expressive, unified, and user-centric systems.

\stitle{Systems for GRMs.}
Recently, there has been a continuous emergence of generative recommendation systems in the industry, covering both training and inference optimizations~\cite{deng2025onerec, xu2025climber, han2025mtgr, huang2025towards, zhou2025onerec, luo2025qarm, wang2025mtgenrecefficientdistributedtraining, cao2025onepiece, dong2025scaling, wang2025scaling, zhu2025llm, dai2025onepiece, yang2024unifying, paischer2024preference, yin2024unleash, penha2024bridging, li2024calrec, ding2024inductive, kim2024sc, shao2024ulmrec, lin2026token}. For instance, OneRec~\cite{deng2025onerec} focuses on single-stage generative recommendation by adopting an MoE architecture and an Iterative Preference Alignment strategy. MTGenRec~\cite{wang2025mtgenrecefficientdistributedtraining} leverages dynamic embedding tables, automated table merging, and two-stage ID deduplication to enhance the scalability of industrial models. Unlike these previous systems, \method identifies the unique optimization potential in the user-side KV cache and employs hierarchical storage and asynchronous pipelining to effectively improve inference performance.

\end{document}